%% file: acl_latex.tex
\pdfoutput=1

\documentclass[11pt]{article}

\usepackage[final]{acl}

\usepackage{times}
\usepackage{latexsym}

\usepackage[T1]{fontenc}

\usepackage[utf8]{inputenc}

\usepackage{microtype}

\usepackage{inconsolata}

\usepackage{graphicx}
\usepackage{amsmath}
\usepackage{amsfonts}
\usepackage[linesnumbered,ruled]{algorithm2e}

\usepackage{xcolor}
\usepackage{svg}

\usepackage{float}
\usepackage{stfloats}

\usepackage{adjustbox}

\usepackage{colortbl}
\usepackage{booktabs}

\newcommand{\midsepremove}{\aboverulesep = 0mm \belowrulesep = 0mm}
\midsepremove
\newcommand{\midsepdefault}{\aboverulesep = 0.605mm \belowrulesep = 0.984mm}
\midsepdefault

\usepackage{multirow}

\usepackage{tabularx}

\usepackage{tcolorbox}

\tcbuselibrary{breakable}
\usepackage{listings}

\usepackage{enumitem}

\usepackage{array}

\title{Distill-C: Enhanced NL2SQL via Distilled Customization with LLMs}

        \author{Cong Duy Vu Hoang$^{1*}$, Gioacchino Tangari$^{2*}$, Clemence Lanfranchi$^{3*}$, \\
        {\bf Dalu Guo$^2$, Paul Cayet$^3$, Steve Siu$^2$, Don Dharmasiri$^2$, Yuan-Fang Li$^2$,} \\
        {\bf Long Duong$^2$, Damien Hilloulin$^3$, Rhicheek Patra$^3$, Sungpack Hong$^3$, Hassan Chafi$^3$} \\
        $^1$Oracle Analytics Cloud (OAC), Australia \\
        $^2$Oracle Health \& AI (OHAI), Australia \\
        $^3$Oracle Labs, Switzerland \\
        \texttt{\{vu.hoang, gioacchino.tangari, clemence.lanfranchi\}@oracle.com}}

\begin{document}

\maketitle

\def\thefootnote{*}\footnotetext{Equal contributions \& corresponding authors}\def\thefootnote{\arabic{footnote}}

\begin{abstract}
\input{sections/sec_abs}
\end{abstract}

\input{sections/sec_intro}

\input{sections/sec_method}

\input{sections/sec_exp}

\input{sections/sec_related_work}

\input{sections/sec_conclusion}

\input{sections/sec_ack}

\bibliography{custom}

\clearpage
\appendix
\input{sections/sec_appendix}

\end{document}

%% file: sections/sec_abs.tex
The growing adoption of large language models (LLMs) in business applications has amplified interest in Natural Language to SQL (NL2SQL) solutions, in which there is competing demand for high performance and efficiency. Domain- and customer-specific requirements further complicate the problem. To address this conundrum, we introduce Distill-C, a distilled customization framework tailored for NL2SQL tasks. Distill-C utilizes large teacher LLMs to produce high-quality synthetic data through a robust and scalable pipeline. Finetuning smaller and open-source LLMs on this synthesized data enables them to rival or outperform teacher models an order of magnitude larger. 
Evaluated on multiple challenging benchmarks,\footnote{Datasets are available at \url{https://github.com/ClemenceLanfranchi/Distill-C}} Distill-C achieves an average improvement of 36\% in execution accuracy compared to the base models from three distinct LLM families. 
Additionally, on three internal customer benchmarks, Distill-C demonstrates a 22.6\% performance improvement over the base models. 
Our results demonstrate that Distill-C is an effective, high-performing and generalizable approach for deploying lightweight yet powerful NL2SQL models, delivering exceptional accuracies while maintaining low computational cost.

%% file: sections/sec_intro.tex
\section{Introduction}
The increasing capabilities of large language models (LLMs) have led to their growing integration into business environments for streamlining routine tasks \cite{minaee2024largelanguagemodelssurvey,liu2024surveynl2sqllargelanguage}. 
A key application is NL2SQL (Natural Language to SQL) translation, where developers frequently need to generate SQL queries for diverse business use cases \cite{zhu2024largelanguagemodelenhanced}. 
Although state-of-the-art LLMs achieve high performance on public benchmarks, their large resource and computational demands, coupled with performance limitations in certain real-world contexts, make smaller specialized models a more suitable option for many practical applications. 
However, smaller LLMs often underperform relative to their larger counterparts, limiting their practical effectiveness in demanding scenarios.

One of the primary motivations for this work is the emerging area of NL2SQL data synthesis and knowledge distillation. 
Existing research has explored approaches to data synthesis and distillation for NL2SQL applications, yet these methods remain generalized rather than tailored to the specific needs of real-world customer environments. 
In recent work \cite{yang-etal-2024-synthesizing} propose a "SQLer" model that generates training examples across diverse topics and domains. However, this approach does not tailor the distillation process to specific business applications.
Similarly, another study \cite{chen-etal-2023-personalized} introduced personalized distillation for code generation by addressing small-model code execution errors, though it is not extended to NL2SQL.

We propose Distill-C (\textbf{Distill}ed \textbf{C}ustomization), a novel framework for NL2SQL distillation that introduces customizable elements to address specific customer use cases, requirements, and expectations. 
Distill-C leverages teacher LLMs to generate distilled knowledge, which is then transferred to smaller student models. 
By incorporating customized synthesis techniques, error-driven reference examples, and tailored distillation strategies, our approach enhances the accuracy and resource efficiency of smaller NL2SQL models, making them more practical for real-world applications. 

Our contributions feature a scalable pipeline with the following key components:
\begin{itemize}[noitemsep,topsep=0pt]
    \item \textbf{Customization}: Integrates customer-specific features into the data synthesis for high-quality NL2SQL data.
    \item \textbf{Targeted Distillation}: Utilizes an ensemble of LLMs to balance their strengths and weaknesses, generating tailored datasets with features like date-time handling, financial analytics, and SQL compliance.
    \item \textbf{Modular Synthesis}: Separates natural language and SQL synthesis, leveraging multiple LLMs for better data diversity and robustness.
    \item \textbf{Quality Assurance}: Uses a multi-step filtering process (pattern matching, execution checks, LLM juries) to refine data quality.
\end{itemize}

Our Distill-C framework effectively enables small LLMs to perform on par with, or even surpass, their teacher models, exhibiting  gains of 36\% on average across different families of models and on various challenging benchmarks.

%% file: sections/sec_method.tex
\begin{figure*}[t!]
    \centering
    \includegraphics[width=\textwidth]{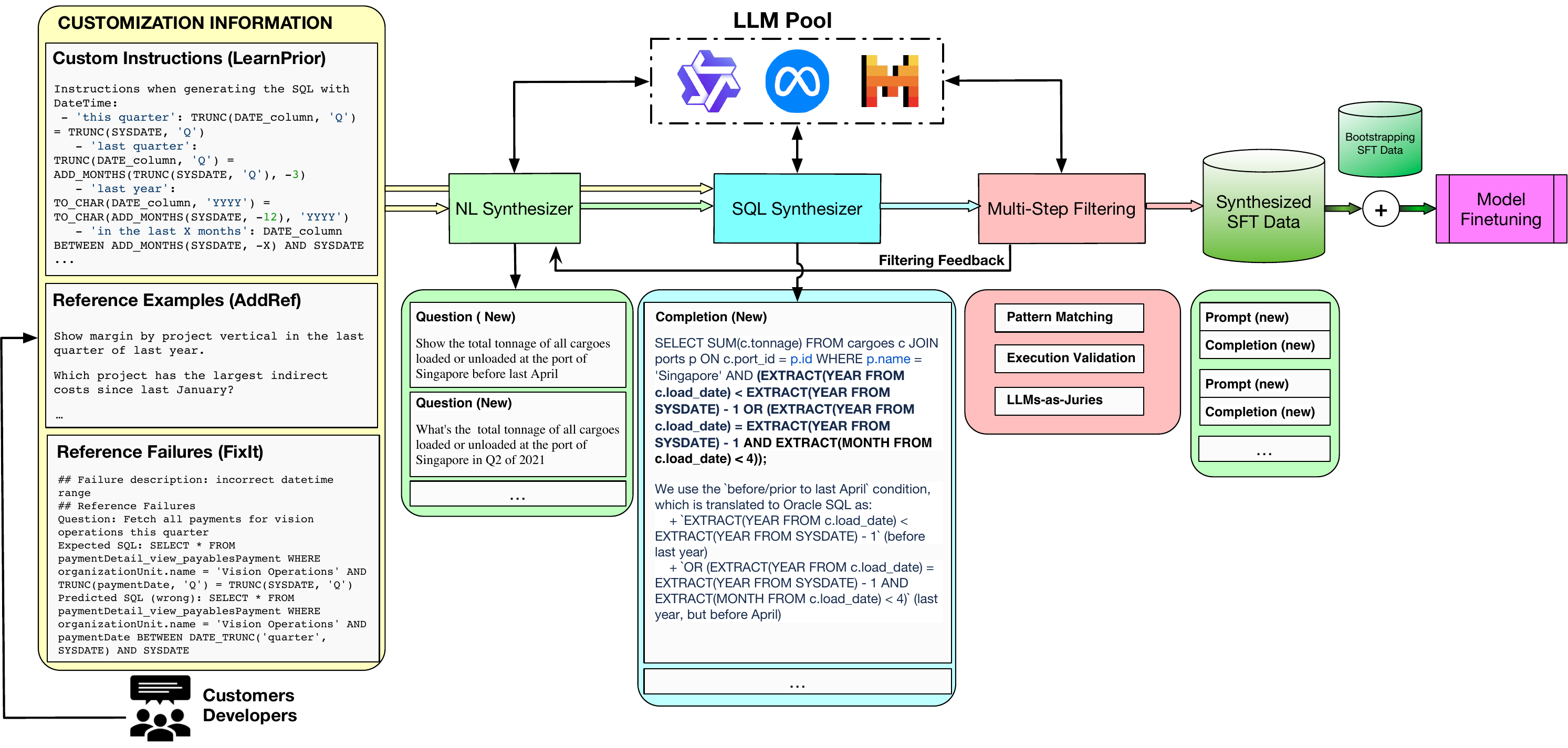}  
    \caption{\textbf{The Proposed Distill-C Framework.}}
    \label{fig:distillc_framework}
\end{figure*}

\section{Methodology}
\subsection{Customization Scenarios}
We present three distinct scenarios, including \textbf{AddRef}, \textbf{LearnPrior}, and \textbf{FixIt} - each of which is based on a reasonable assumption often confirmed in enterprise settings, where product and engineering teams typically have the capacity to provide a few examples, instructional guidance, and error feedback from early model deployments.

\paragraph*{AddRef: Incorporating Reference Examples.} Reference examples consist of a pre-defined subset of natural language (NL) queries provided by the \textbf{Customer} and serve as a basis for guiding data generation by LLMs. It is essential that these generated NL examples not only closely resemble the reference examples but also exceed them in complexity and originality.

\paragraph*{LearnPrior: Leveraging Prior Custom Instructions.} The \textbf{Customer} provides a limited set of statements detailing prior requirements and expectations for SQL responses generated by NL2SQL models. These statements convey the \textbf{Customer's} insights into how model outputs should align with their specific needs.

\paragraph*{FixIt: Distilling Targeted Knowledge from Error Feedback.} In this scenario, the \textbf{Customer} has initial access to a baseline model that is evaluated to identify a set of unacceptable model errors. These errors serve as starting points for bootstrapping targeted improvements, helping the model avoid similar issues in subsequent iterations.

\subsection{The Distill-C Framework}
We developed our \textbf{Distill}ed \textbf{C}ustomization framework, abbreviated as \textbf{Distill-C}, to synthesize tailored knowledge specifically adapted to the customer scenarios described above. 
The core components of our proposed \textbf{Distill-C} framework are illustrated in Figure~\ref{fig:distillc_framework}. The framework comprises distinct NL and SQL synthesizers, followed by a three-stage filtering pipeline, and it enables the integration of knowledge from multiple advanced LLMs at each stage.

\begin{figure*}[h!]
    \centering
    \includegraphics[width=\textwidth]{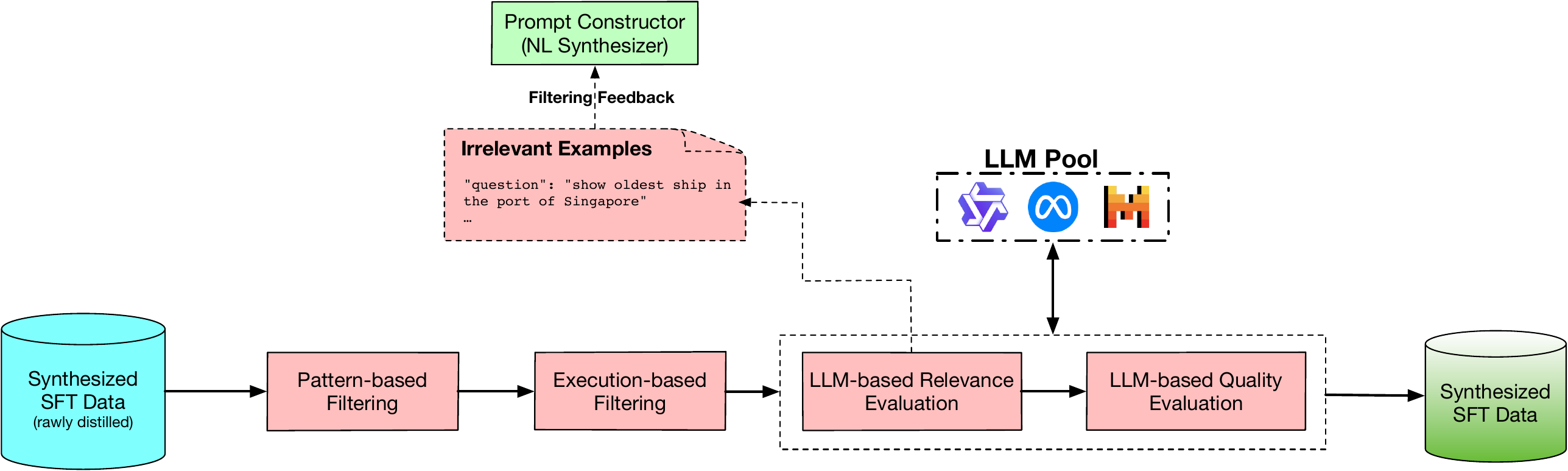}  
    \caption{The Multi-Step Filtering Pipeline in our \textbf{Distill-C} Framework.}
    \label{fig:distillc_multistep_filtering}
\end{figure*}

\subsubsection{Distillation Pipelines}
Our framework decouples NL and SQL synthesis, which, though less resource-efficient than a single-step approach, offers two key benefits: \textit{First}, independent generation by different LLMs enhances data diversity; \textit{Second}, it leverages model-specific strengths. For example, while Llama3.1-70B-Instruct excels at generating realistic queries for a database schema, it may miss OracleSQL-specific nuances better addressed by Mixtral-8x22B-Instruct-v0.1, as shown in Table~\ref{tab:full_perf}.

\paragraph*{NL Synthesizer Pipeline.} The NL synthesizer produces new NL queries or questions, guided by the customer's customization scenarios, including reference NL examples\footnote{consisting of 100 examples or fewer to initiate the data synthesis process.} (AddRef); prior expert instructions (LearnPrior); and targeted knowledge from error feedback (FixIt). These scenarios can be applied individually or in combination.

The NL synthesis process\footnote{as further illustrated in Appendix Figure~\ref{fig:distillc_nl_synthesis}.} begins with \textbf{Reference NL Extraction \& Sampler}, where NL queries are sampled from reference examples, balancing inspiration with diversity within the LLM’s context window. 
The \textbf{Prompt Constructor} then assembles NL generator prompts by combining these sampled NL examples and a database (DB) schema.\footnote{sampled from a pool of training DB schemas.} 
We also utilize discarded examples from previous generation rounds, incorporating a limited selection of them into the prompt as negative examples, which helps to iteratively refine the natural language synthesis process. 

Finally, multiple LLMs (preferably 50B+ parameters) generate diverse NL queries by leveraging high-temperature sampling and varied random seeds, benefiting from their superior instruction following and generation diversity.\footnote{Despite their capability, proprietary LLMs (OpenAI; Anthropic; Gemini) are excluded from this process due to licensing restrictions on production use of their generated data.} 
The outcome of the NL synthesis phase is a set of new NL queries relevant to the customer use case, each mapped to a DB schema.

\paragraph*{SQL Synthesizer Pipeline.} Starting with a set of \{NL question, DB schema\} pairs generated in earlier steps, the SQL synthesizer employs multiple Generator LLMs to translate each question into its corresponding SQL query. 
This process produces a preliminary, or "raw", distillation dataset (prior to filtering), where each entry forms a complete NL2SQL data point pairing the DB schema and NL question as the prompt with the SQL query as the completion. 
This dataset serves as a foundation for transferring knowledge from strong foundational LLMs into smaller models.

Key aspects of the SQL synthesis process (detailed in Appendix Figure~\ref{fig:distillc_sql_synthesis}) include:
\begin{itemize}[noitemsep,topsep=0pt]
	\item \textbf{Diverse LLMs as Generators}: 
	Multiple LLMs enhance data diversity and address model-specific gaps, with some excelling in constructs like the Oracle SQL dialect.
	\item \textbf{Instruction-Conditioned Generation}: 
	Task-specific instructions (LearnPrior) ensure SQL outputs align with customer requirements, including handling complex datetime structures (intervals, absolute and relative references).\footnote{as further illustrated in Appendix Figure~\ref{fig:prompt_sql_syn}.}
\end{itemize}
The synthesis process includes three key steps:
\begin{enumerate}[noitemsep,topsep=0pt]
    \item \textbf{Prompt Constructor}: 
    Combines user queries, database schemas, and task-specific instructions to create effective prompts.
	\item \textbf{SQL Generation}: 
	LLMs generate SQL queries with descriptions, forming a synthetic supervised fine-tuning (SFT) dataset that clarifies complex SQL elements.
	\item \textbf{Prompt Post-Processing}: 
	Strips instructions from prompts in the SFT dataset to ensure smaller models learn directly from distilled examples.
\end{enumerate}

\subsubsection{Multi-Step Filtering Pipeline}
The training examples derived from the NL \& SQL Synthesizer pipelines, consisting of (i) a prompt with a new question and (ii) an SQL completion, undergo a multi-step filtering process, as illustrated in Figure~\ref{fig:distillc_multistep_filtering}, to ensure data quality and minimize noise:
\begin{itemize}[noitemsep,topsep=0pt]
    \item \textbf{Pattern-Based Filtering}: Removes examples with non-target syntax (e.g., MySQL-specific keywords for Oracle SQL), reducing the load on resource-intensive downstream filters.
    \item \textbf{Execution-Based Filtering}: Validates SQL by executing it on real databases linked to schema contexts, discarding non-executable queries to prevent negatively impacting model performance.
    \item \textbf{LLM-Based Quality Evaluation}: Uses multiple strong LLMs as "juries" \cite{verga2024replacingjudgesjuriesevaluating} to evaluate and rank examples for semantic accuracy to ensure alignment with intended NL meaning. This automated approach replaces manual review for large datasets.
    \item \textbf{LLM-Based Relevance Evaluation}: Ensures examples are relevant to the target use case by requiring unanimous agreement among LLMs. Irrelevant data is flagged as "Filtering Feedback" (Figure~\ref{fig:distillc_framework}) for refining the NL synthesis.
\end{itemize}

\subsubsection{Finetuning}
The final step involves finetuning the smaller target LLM using synthesized instruction data and a small bootstrapping dataset, which is crucial for mitigating biases and preventing model collapse \cite{gerstgrasser2024modelcollapseinevitablebreaking}.

%% file: sections/sec_exp.tex
\begin{table*}[ht]
\centering
\scriptsize
\begin{adjustbox}{width=\textwidth}
\midsepremove
\begin{tabular}{lcccccccc}
\toprule
\multirow{2}{*}{\textbf{Model Variant}} & \multicolumn{4}{c}{\textbf{DateTime (\%)}} & \multicolumn{2}{c}{\textbf{Financial Analytics (\%)}} & \multicolumn{2}{c}{\textbf{\begin{tabular}[c]{@{}c@{}}OracleSQL\\ Compliance (\%)\end{tabular}}} \\ \cmidrule(lr){2-5} \cmidrule(lr){6-7} \cmidrule(lr){8-9}
 & \textbf{spd-ora} & \textbf{spd-lite} & \textbf{bd-lite} & \textbf{bd-ora} & \textbf{spd+bd-ora} & \textbf{spd+bd-lite} & \textbf{spd-ora} & \textbf{bd-ora} \\ \midrule
\multicolumn{9}{c}{\textbf{Student LLMs \& SFT with Distill-C (A-Full Setting)}} \\ \midrule
\rowcolor[HTML]{FFF9E3}
CodeQwen1.5-7B-Chat & 30.4 & 58.1 & 37.9 & 2.6 & 24.8 & 47.8 & 33.9 & 4.6 \\
\rowcolor[HTML]{E8F8E8} 
+Distill-C (A-Full) & \textbf{74.0} & \textbf{68.7} & \textbf{57.2} & \textbf{33.8}\dag & \textbf{89.5}\dag & \textbf{84.1}\dag & \textbf{77.6} & \textbf{34.8}\dag \\ \hline
\rowcolor[HTML]{FFF9E3}
Llama-3.1-8B-Instruct & 29.8 & 62.6 & 41.3 & 2.6 & 17.0 & 35.9 & 36.1 & 3.1 \\
\rowcolor[HTML]{E8F8E8} 
+Distill-C (A-Full) & \textbf{81.2}\dag & \textbf{67.6} & \textbf{59.3} & \textbf{29.5} & \textbf{83.2} & \textbf{78.2} & \textbf{79.4}\dag & \textbf{32.0} \\ \hline
\rowcolor[HTML]{FFF9E3} 
Mistral-7B-Instruct-v0.3 & 22.1 & 46.4 & 22.2 & 2.6 & 21.1 & 24.5 & 38.4 & 4.4 \\
\rowcolor[HTML]{E8F8E8} 
+Distill-C (A-Full) & \textbf{74.6} & \textbf{65.4} & \textbf{38.8} & \textbf{31.2} & \textbf{84.5} & \textbf{80.4} & \textbf{77.3} & \textbf{28.2} \\ \midrule
\multicolumn{9}{c}{\textbf{Out-of-the-Box Strong LLMs (selected)}} \\ \midrule
Qwen2-72B-Instruct (teacher) & 32.0 & 67.0 & 55.7 & 8.1 & 41.2 & 62.1 & 42.4 & 9.0 \\
Llama-3.1-70B-Instruct (teacher) & 24.3 & 62.0 & \textbf{61.6}\dag & 4.3 & 1.6 & 42.3 & 34.4 & 4.4 \\
Mixtral-8x22B-Instruct-v0.1 (teacher) & 48.6 & 64.8 & 42.0 & 21.4 & 67.5 & 71.3 & 54.1 & 16.9 \\
Mistral-Large-Instruct-2407 & 51.4 & \textbf{73.7}\dag & 53.9 & 16.2 & 83.6 & 83.2 & 58.1 & 20.6 \\
DeepSeek-Coder-V2-Instruct & 44.2 & 71.5 & 55.3 & 15.0 & 65.2 & 78.2 & 53.8 & 19.4 \\ \bottomrule
\end{tabular}
\end{adjustbox}
\caption{\textbf{Task performances on DateTime, Financial Analytics, and OracleSQL Compliance.} \dag marks column bests; bold shows Distill-C induced performance. Notations: spd: Spider, bd: Bird, ora: OracleSQL, lite: SQLite.}
\label{tab:full_perf}
\end{table*}
\begin{table*}[ht]
\centering
\begin{adjustbox}{width=\textwidth}
\begin{tabular}{p{2cm}p{7.5cm}p{3cm}p{3cm}p{3.8cm}}
\toprule
\textbf{Customer} & \textbf{Use Case} & \textbf{Student Model} & \textbf{Distill-C Model} & \textbf{Distill-C Impact} \\ 
\midrule
\textbf{Customer 1} & Account payables and receivables management (4 schemas; 192/497 examples with datetime) & 80\%         & \textbf{97\%}         & Distill-C $\rightarrow$ DateTime       \\
\textbf{Customer 2} & Information technology services and consulting (1 schema; 25/28 examples with financial analytics) & 54\%         & \textbf{78\%}         & Distill-C $\rightarrow$ Financial Analytics \\ 
\textbf{Customer 3} & Autonomous database (6 schemas; 99/99 examples with OracleSQL compliance) & 42\%         & \textbf{71\%}         & Distill-C $\rightarrow$ OracleSQL Compliance \\ \bottomrule
\end{tabular}
\end{adjustbox}
\caption{Impact of Our Distill-C Method on Customer Benchmarks.}
\label{tab:business_impact}
\end{table*}

\begin{table}[t!]
\centering
\scriptsize
\begin{adjustbox}{width=\columnwidth}
\midsepremove
\begin{tabular}{llccccc}
\toprule
\textbf{Task} & \textbf{Origin} & \textbf{SQL Dialect} & \textbf{Train} & \textbf{Dev} & \textbf{Test} \\ \midrule
\textbf{DateTime}      & Bird   & OracleSQL & 9,621 & 115 & 533 \\ 
              & Bird   & SQLite    & 33,173 & 78 & 234 \\ 
              & Spider & OracleSQL & 13,460 & 131 & 680 \\ 
              & Spider & SQLite    & 37,172 & 97 & 179 \\ 
\midrule
\textbf{Financial} & Bird   & OracleSQL & 13,460 & 63 & 1,753 \\ 
\textbf{Analytics} & Bird   & SQLite    & 23,091 & 113 & 734 \\ 
              & Spider & OracleSQL & 17,734 & 108 & 3,820 \\ 
              & Spider & SQLite    & 35,749 & 123 & 1,366 \\ 
\midrule
\textbf{OracleSQL} & Bird & OracleSQL & 29,877 & 319 & 1,469 \\ 
\textbf{Compliance} & Spider & OracleSQL & 39,369 & 326 & 1,478 \\ 
\bottomrule
\end{tabular}
\end{adjustbox}
\caption{Statistics of Train, Dev, and Test Datasets.}
\label{tab:data_stats}
\end{table}

\section{Experiments}
\subsection{Evaluation Tasks}
We evaluate our approach on customer-identified tasks, including:
\begin{itemize}[noitemsep,topsep=0pt]
    \item \textbf{DateTime}: Generating SQL for complex temporal conditions, including relative (e.g., \textit{"last 2 quarters"}) and composite clauses (e.g., \textit{"first quarter of the last 5 years"}).
    \item \textbf{Financial Analytics}: Querying trends, correlations, and financial metric breakdowns (e.g., profits by country or quarter).
    \item \textbf{OracleSQL Compliance}: Producing syntactically correct OracleSQL queries.
\end{itemize}

\subsection{Data and Evaluation Settings}
\textbf{Experimental Data.}
We built our experimental data using Spider (1.0) \cite{yu2018spider} and BIRD \cite{bird_10.5555_3666122.3667957}. For each task, we prepared three datasets: (i) a curated test set; (ii) a small development set for customization via AddRef, LearnPrior, and FixIt scenarios; (iii) a training set generated with the \textbf{Distill-C} pipeline. The training, testing and dev sets respectively comprise 199, 31, 10  disjoint DB schemas to prevent data leakage. Data statistics are in Table~\ref{tab:data_stats}.
\begin{table}[t!]
\scriptsize
\centering
\midsepremove
\begin{tabularx}{\columnwidth}{cX}
\toprule
\textbf{Setting} & \textbf{Description} \\ \midrule
\multirow{2}{*}{\textbf{B}} & Distill-C w/ AddRef (NL): Uses 10 to 100 NL-only examples for data synthesis without SQL supervision. \\ \midrule
\multirow{2}{*}{\textbf{C}} & Distill-C w/ AddRef (NL) + LearnPrior: Adds tailored instructions to NL-only examples to guide SQL generation. \\ \midrule
\multirow{2}{*}{\textbf{D}} & Distill-C w/ AddRef (NL+SQL): Adds SQL supervision with paired NL + SQL examples for explicit NL-to-SQL mappings. \\ \midrule
\multirow{2}{*}{\textbf{E}} & Distill-C w/ AddRef (NL) + LearnPrior + FixIt: Extends \textbf{C} with incorrect SQL examples to train error recognition. \\ \midrule
\multirow{1}{*}{\textbf{A-Full}} & Full Distill-C: AddRef (NL+SQL) + LearnPrior + FixIt \\ \bottomrule
\end{tabularx}
\caption{Summary of evaluation settings.}
\label{tab:settings}
\end{table}

\noindent\textbf{Metric.} We use execution accuracy \cite{zhong-etal-2020-semantic} to evaluate our framework, which compares the execution results of the generated SQL query and the ground-truth on the corresponding database.

\subsection{Model Settings}
We evaluated our proposed Distill-C framework with a series of settings, progressing from NL-only (B) to complete (A-Full), which enables systematic evaluation of the impact of increasing supervision and tailored training signals on model performance, as shown in Table~\ref{tab:settings}. The distillation signals from teacher LLMs are derived in Table~\ref{tab:distillation_signals}.
\begin{table}[b!]
\scriptsize
\centering
\midsepremove
\begin{tabularx}{\columnwidth}{lX}
\toprule
\textbf{Student LLM} & \textbf{Teacher LLM(s)} \\ \midrule
Qwen1.5-7B-Instruct & Qwen2-72B-Instruct, Mixtral-8x22B-Instruct-v0.1 \\ 
Llama3.1-8B-Instruct & Llama3.1-70B-Instruct, Mixtral-8x22B-Instruct-v0.1 \\ 
Mistral-7B-Instruct-v0.3 & Mixtral-8x22B-Instruct-v0.1, Llama3.1-70B-Instruct \\ \bottomrule
\end{tabularx}
\caption{Student \& Teacher LLMs used for distillation.}
\label{tab:distillation_signals}
\end{table}

\subsection{Public Main Results}
The experimental results in Table \ref{tab:full_perf} highlight the effectiveness of our proposed \textbf{Distill-C} framework, which integrates three customization scenarios (AddRef, LearnPrior, FixIt) to enhance the performance of various student LLMs across three challenging tasks: DateTime, Financial Analytics, and Oracle SQL Compliance.
Our approach achieves significant performance gains across three foundational LLMs: CodeQwen1.5-7B-Chat (26.2\%, 55.5\%, 36.9\%), Llama-3.1-8B-Instruct (25.3\%, 54.3\%, 36.1\%), and Mistral-7B-Instruct-v0.3 (29.2\%, 59.7\%, 31.4\%) for DateTime, Financial Analytics, and OracleSQL Compliance, respectively. These improvements across multiple benchmarks underscore the robustness of our method in enhancing LLM capabilities across diverse tasks. 
\begin{figure*}[t!]
    \centering
    \includegraphics[width=0.90\textwidth]{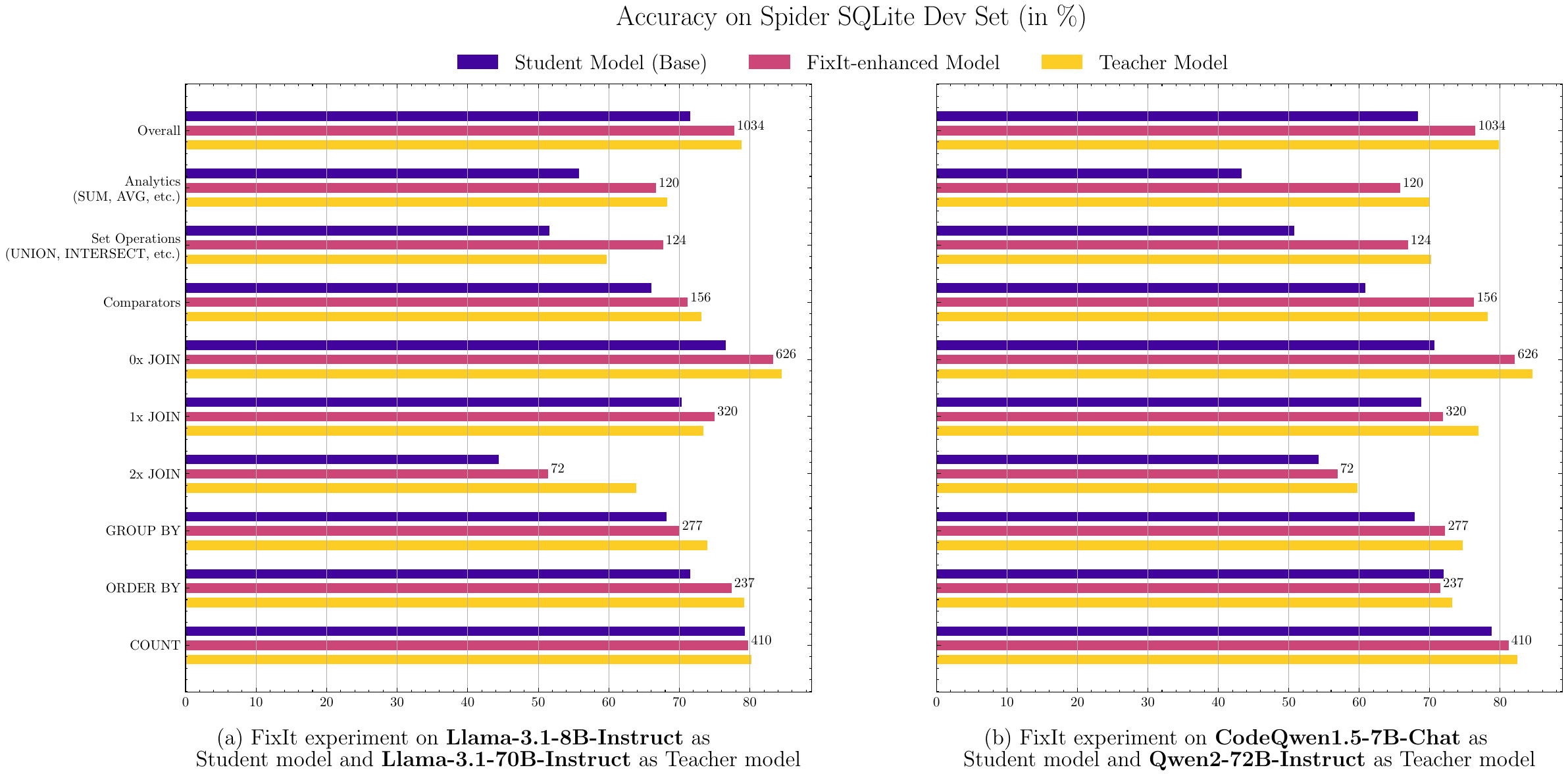}  
	\caption{\textbf{FixIt Ablation Study Experiments.} Performance of student models finetuned with the FixIt scenario using Distill-C on Spider (dev) sub-groups, showing results for student, finetuned, and teacher models, with sample counts per group.}
    \label{fig:fixit_exp}
\end{figure*}

Furthermore, the distilled models surpass several strong out-of-the-box LLMs, including their teacher models such as Qwen2-72B-Instruct, Llama-3.1-70B-Instruct, and Mixtral-8x22b-Instruct-v0.1, which can be attributed to the tailored prompts that are used to guide the data synthesis process, fostering better SQL generation from the teacher models. 
Our fine-tuned models outperform larger state-of-the-art LLMs (e.g., Mistral-Large-Instruct-2407 and DeepSeek-Coder-V2-Instruct) on multiple benchmarks, showcasing the effectiveness of the Distill-C framework. 
These findings demonstrate the potential of the Distill-C framework to significantly enhance smaller LLMs, enabling them to handle complex tasks more effectively while providing substantial efficiency benefits for deployment.

\subsection{Customer Impact}
We demonstrated the business impact of our Distill-C method through enhanced performance gains on internal and customer-specific datasets.\footnote{Due to proprietary restrictions, we are unable to disclose the specifics of the customer schemas as well as benchmark sets for the NL2SQL tasks.}

The performance boost of Distill-C on domain-specific tasks, as shown in Table~\ref{tab:business_impact}, highlights its capability to address key challenges in customer-specific tasks such as DateTime handling, financial analytics, and SQL compliance, improving average accuracy significantly, by 22.6 absolute points. 
For DateTime tasks in Customer 1's account management use case, Distill-C achieved near-perfect accuracy (97\%), demonstrating its robustness in handling temporal data critical for financial workflows. 
In Customer 2's financial analytics use case, the model significantly improved performance from 54\% to 78\%, showcasing its ability to handle complex financial datasets and provide actionable insights. 
Finally, for Customer 3, focused on OracleSQL compliance in autonomous database use case, Distill-C delivered a substantial gain, raising accuracy from 42\% to 71\%. 
These results underscore Distill-C's versatility and effectiveness in enhancing precision and reliability across specialized tasks in diverse domains.

\subsection{Ablation Study}
We conduct two ablation studies to assess the impact of individual scenarios in Distill-C.

\textbf{Individual FixIt Scenario.} We evaluate the FixIt scenario using Llama-3.1-8B-Instruct and CodeQwen1.5-7B-Chat as student LLMs. Errors identified from the Spider training set \cite{yu2018spider} are processed through our data generation pipeline, where the NL Prompt Constructor (Figure~\ref{fig:distillc_nl_synthesis}) utilize these errors to guide teacher LLMs (Qwen2-72B-Instruct for CodeQwen and Llama-3.1-70B-Instruct for Llama) to create targeted datasets used to finetune the student models, producing FixIt-enhanced versions.

On the Spider development set, FixIt achieves performance improvements of 6.4\% and 8\% for Llama-3.1-8B-Instruct and CodeQwen1.5-7B-Chat, respectively, significantly narrowing gaps with their teacher models. Figure~\ref{fig:fixit_exp} shows notable gains in Analytics and Set Operations, effectively addressing key weaknesses.

\begin{figure*}[t!]
    \centering
    \includegraphics[width=\textwidth]{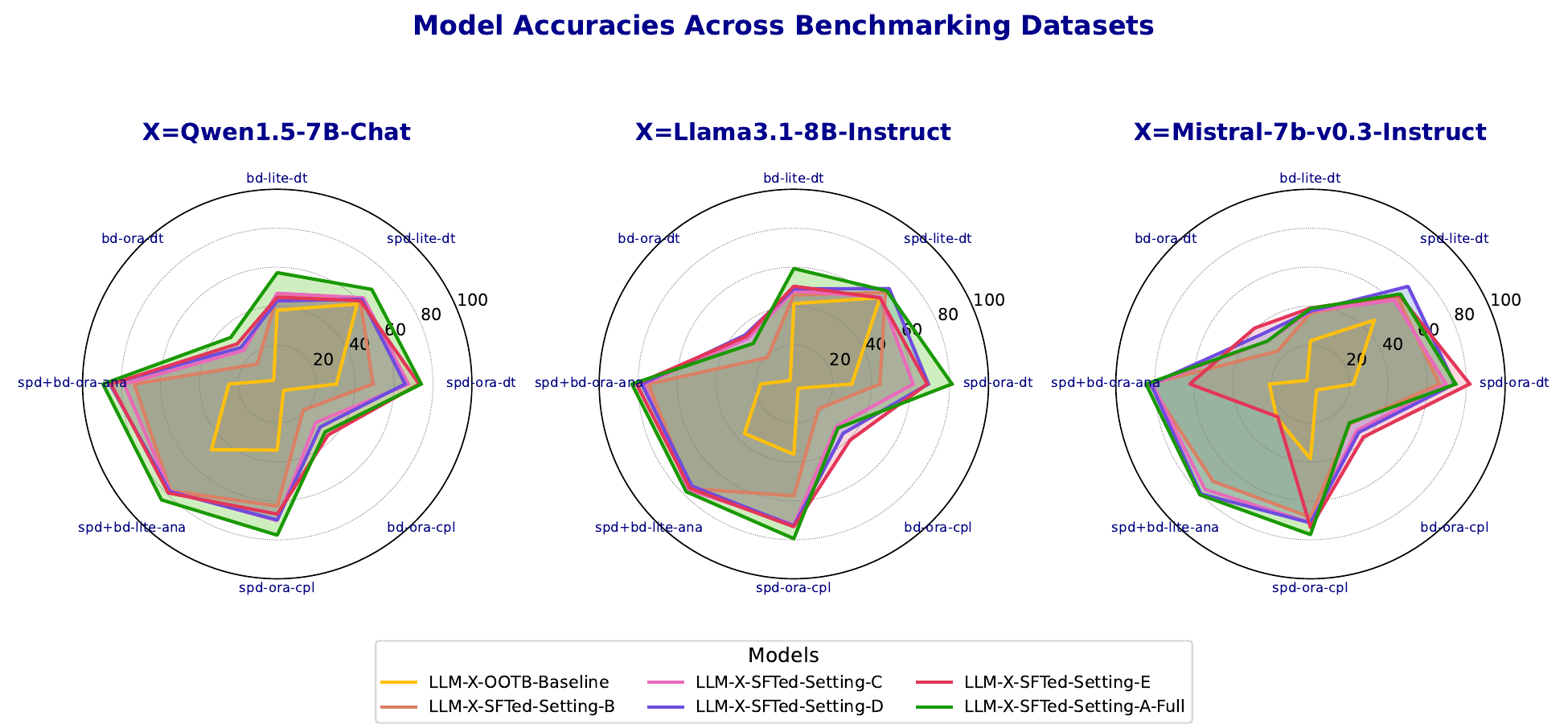}
    \caption{\textbf{Ablation study with distillation settings (Table~\ref{tab:settings}).} Notations: spd: Spider, bd: Bird, dt: DateTime, ana: Analytics, ora: OracleSQL, lite: SQLite, cpl: Compliance. Numerical results are reported in Appendix~\ref{appendix:abl}.}
    \label{fig:full_ablation_exp}
\end{figure*}

\textbf{Full Scenarios.} 
Figure~\ref{fig:full_ablation_exp} demonstrates the consistent and substantial improvements achieved by integrating all scenarios (AddRef+LearnPrior+FixIt) within our Distill-C framework. While the AddRef scenario alone (Setting B) already brings a significant improvement of 24.7\% on average, showcasing the importance of finetuning models on tasks that are similar to the target tasks, we also see that providing prior knowledge and leveraging errors is key to obtaining optimal performance. Moreover, the similarity in performance between scenarios C and D (respectively +30.4\% and +32.6\% on average) tends to show that custom instructions and examples of ground truth SQL queries are both valid options to distill prior knowledge. 
This integration leads to significant performance gains across a diverse range of benchmarks, including DateTime, Financial Analytics, and Oracle SQL Compliance, showcasing the versatility and robustness of our approach. Notably, these improvements are observed consistently across multiple student LLMs, underscoring the generalizability and effectiveness of the proposed framework. 
Overall, the results highlight how the synergistic combination of these scenarios enables Distill-C to address complex challenges and deliver superior outcomes, making it a compelling solution for advancing language understanding and task-specific performance.


%% file: sections/sec_related_work.tex

\section{Related Work}
Recent advancements in NL2SQL research have explored techniques to enhance the performance of Large Language Models (LLMs).

\textbf{Prompt Engineering and Reasoning.} Prompt engineering has been explored to optimize NL2SQL capabilities of LLMs. PET-SQL \cite{li2024petsqlpromptenhancedtworoundrefinement} adopts a two-round framework with enhanced representations, and EPI-SQL \cite{liu2024episqlenhancingtexttosqltranslation} generates error-prevention prompts to reduce LLM errors. Self-correction and iterative refinement have also been explored in SQL-CRAFT \cite{xia2024r3thissqlme} and DART-SQL \cite{mao-etal-2024-enhancing}, which integrate interactive feedback loops. 
However, these approaches are not well-suited to smaller Large Language Models (LLMs) because they necessitate acute reasoning capabilities that such models typically lack. On the other hand, Distill-C addresses this limitation by focusing on bridging the performance gap between large and small LLMs. This method leverages the advanced reasoning abilities of larger LLMs to distill their knowledge into more compact forms, thereby enhancing the capabilities of smaller models without requiring extensive computational resources.


\textbf{Synthetic Data Generation.} Recent works have shown the great promise of synthetic data. SQL-GEN \cite{pourreza2024sqlgenbridgingdialectgap} produces dialect-specific synthetic training data, while SENSE \cite{yang2024synthesizingtexttosqldataweak} utilizes synthetic data for domain generalization and preference learning. 
Our approach focuses on creating tailored datasets that cater to specific customer needs by integrating targeted instructions and relevant examples into our data generation pipeline. 
Unlike previous work, we further customize the data generation process for individual student language models (LLMs) using error-driven reference examples.


%% file: sections/sec_conclusion.tex

\section{Conclusion}

We introduce Distill-C, a novel customizable distillation framework for enhancing small LLMs in NL2SQL tasks for enterprise applications. Despite their smaller sizes, the enhanced models by Distill-C achieve significant gains over strong baselines across benchmarks, including DateTime, Financial Analytics, and Oracle SQL Compliance. 
The initial costs associated with Distill-C, which involve hosting larger LLMs for data generation and fine-tuning smaller models, are offset by long-term advantages. These benefits arise because business units can then utilize more efficient and specialized smaller LLMs, ultimately leading to a substantial return on investment. 
Our work lays the foundation for robust distillation solutions, enabling the development of specialized NL2SQL models that can be tailored to specific business needs.

Our future work will explore extensions to preference alignment training \cite{10.5555/3666122.3668460} and applications to other practical tasks.

%% file: sections/sec_ack.tex
\section*{Acknowledgments}
We extend our sincere appreciation to our colleagues at the Science Team within Oracle Cloud Infrastructure (OCI) for their support and valuable feedback.

We are grateful to Giulia Carocari for her assistance in translating SQL queries between SQLite and Oracle SQL, as well as the anonymous reviewers whose valuable feedback significantly improved this work.

%% file: sections/sec_appendix.tex
\section{Additional Figures}
\begin{figure*}[h!]
    \centering
    \includegraphics[width=\textwidth]{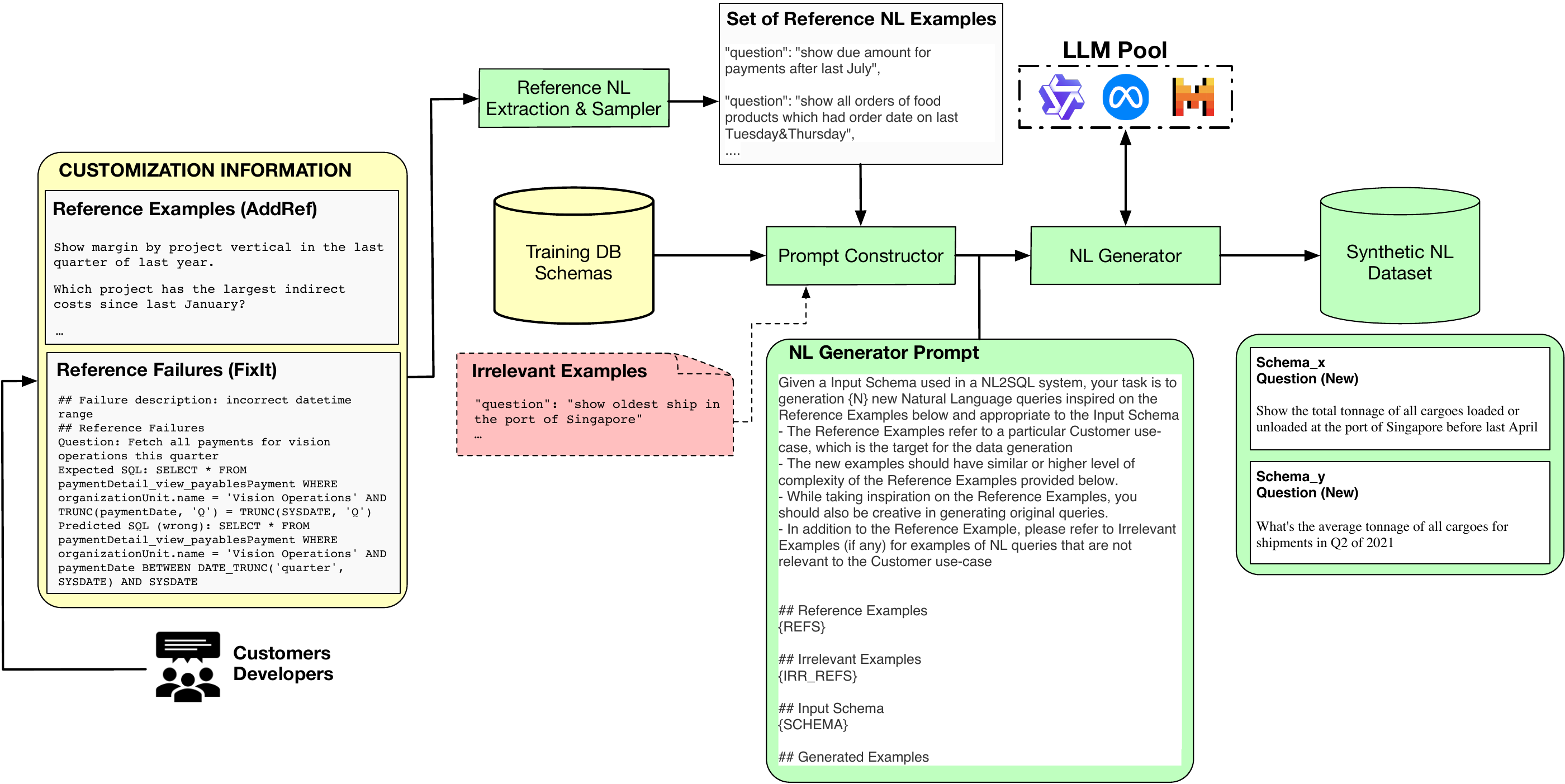}  
    \caption{The NL Synthesizer Pipeline in our \textbf{Distill-C} Framework.}
    \label{fig:distillc_nl_synthesis}
\end{figure*}

\begin{figure*}[h!]
    \centering
    \includegraphics[width=\textwidth]{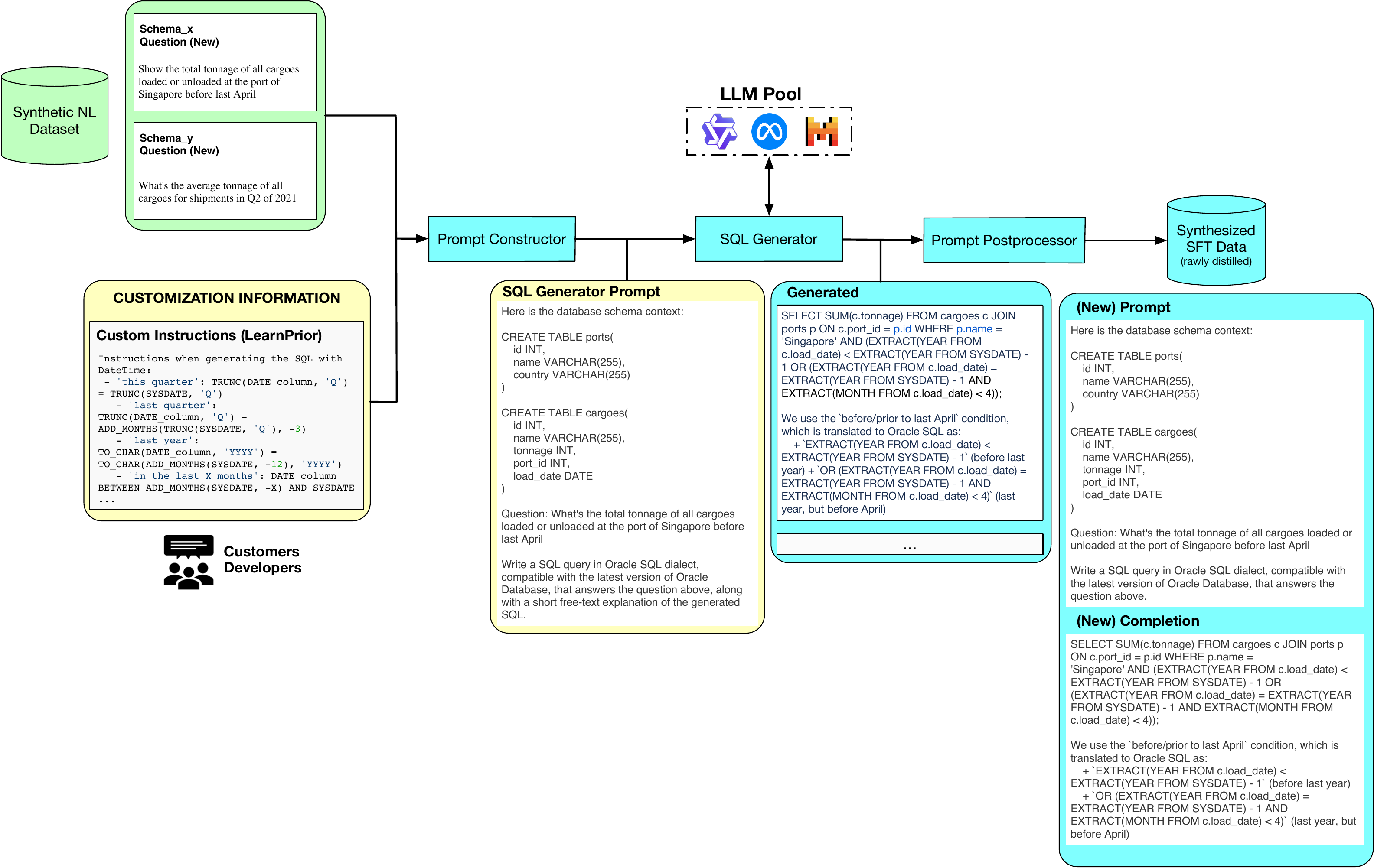}  
    \caption{The SQL Synthesizer Pipeline in our \textbf{Distill-C} Framework.}
    \label{fig:distillc_sql_synthesis}
\end{figure*}

We include additional figures to illustrate the components of our Distill-C framework: the NL Synthesizer Pipeline in Figure~\ref{fig:distillc_nl_synthesis} and the SQL Synthesizer Pipeline in Figure~\ref{fig:distillc_sql_synthesis}, respectively.

\section{Additional Tables}
\subsection{Experimental Setup: Training and Inference Configurations}
We also provide our training and inference hyperparameter configurations in Table~\ref{tab:hparams_config}.

\subsection{Evaluation Tasks}\label{appendix:apptasks}
In Table~\ref{tab:evaltasks}, we present detailed descriptions and examples of the three evaluation tasks used to assess the impact of our \textbf{Distill-C} framework.

\begin{table*}[p!]
\centering
\scriptsize
\begin{tabularx}{\textwidth}{>{\raggedright\arraybackslash}p{0.1\textwidth}X>{\raggedright\arraybackslash}p{0.1\textwidth}>{\raggedright\arraybackslash}p{0.15\textwidth}>{\raggedright\arraybackslash}p{0.30\textwidth}}
\toprule
\textbf{Task Name} & \textbf{Description} & \textbf{DB Schema} & \textbf{Sample NL Query} & \textbf{Sample OracleSQL Query} \\
\midrule

\multirow{2}{=}{\textbf{DateTime}} & \multirow{2}{\linewidth}{Handling complex temporal conditions, including absolute, relative, and composite clauses. Absolute clauses use fixed dates, relative clauses involve SYSDATE, and composite clauses mix both.} & wta\_1 (Spider) & Get the ranking history of Serena Williams since March 2015. & \texttt{SELECT rankings.* FROM rankings JOIN players ON rankings.player\_id = players.player\_id WHERE players.first\_name = 'Serena' AND players.last\_name = 'Williams' AND TO\_CHAR(rankings.ranking\_date, 'YYYY-MM') >= '2015-03'} \\
\cmidrule{3-5}


 &  & financial (Bird) & Which client got his/her card issued since last May? Show the client ID. & \texttt{SELECT T2.client\_id FROM "client" T1 INNER JOIN disp T2 ON T1.client\_id = T2.client\_id INNER JOIN card T3 ON T2.disp\_id = T3.disp\_id WHERE TRUNC(T3.issued, 'MM') >= ADD\_MONTHS(TRUNC(SYSDATE - INTERVAL '1' YEAR, 'YYYY'), 4)} \\
\midrule

\multirow{2}{=}{\textbf{Financial Analytics}} & \multirow{2}{\linewidth}{Producing trends, correlations, and breakdown of financial metrics by date-time intervals and categories. Includes handling complex clauses like GROUP BY, ORDER BY, and Common Table Expressions (CTEs).} & e\_commerce (Spider) & What is the total revenue generated by each product for each customer in 2023, and which product generated the highest revenue for each customer? & \texttt{SELECT c.customer\_id, c.customer\_first\_name, c.customer\_last\_name, p.product\_id, p.product\_name, SUM(p.product\_price) AS total\_revenue, RANK() OVER (PARTITION BY c.customer\_id ORDER BY SUM(p.product\_price) DESC) AS revenue\_rank FROM Customers c JOIN Orders o ON c.customer\_id = o.customer\_id JOIN Order\_Items oi ON o.order\_id = oi.order\_id JOIN Products p ON oi.product\_id = p.product\_id JOIN Shipments s ON o.order\_id = s.order\_id JOIN Invoices i ON s.invoice\_number = i.invoice\_number WHERE EXTRACT(YEAR FROM i.invoice\_date) = 2023 GROUP BY c.customer\_id, c.customer\_first\_name, c.customer\_last\_name, p.product\_id, p.product\_name ORDER BY c.customer\_id, total\_revenue DESC} \\
\cmidrule{3-5}

 &  & financial (Bird) & Calculate the total loans approved per district in 2023, broken down by status, sorted in descending order. & \texttt{SELECT d.district\_id, d.A2 AS district\_name, l.status, SUM(l.amount) AS total\_loan\_amount FROM district d JOIN "account" a ON d.district\_id = a.district\_id JOIN loan l ON a.account\_id = l.account\_id WHERE EXTRACT(YEAR FROM l."date") = 2023 GROUP BY d.district\_id, d.A2, l.status ORDER BY total\_loan\_amount DESC} \\

\midrule

\multirow{2}{=}{\textbf{OracleSQL Compliance}} & \multirow{2}{\linewidth}{Handling OracleSQL-dialect syntax, including ORDER BY with "FETCH FIRST/LAST \{N\} ROWS", correct quoting, and casing for schema object names.} & car\_1 (Spider) & What are the different models created by either General Motors or over 3500 lbs? & \texttt{SELECT DISTINCT T1."model" FROM model\_list T1 JOIN car\_makers T2 ON T1.Maker = T2."id" JOIN car\_names T3 ON T1."model" = T3."model" JOIN cars\_data T4 ON T3.MakeId = T4."id" WHERE T2.FullName = 'General Motors' OR T4.Weight > 3500} \\
\cmidrule{3-5}


 & & financial (Bird) & List out the accounts who have the earliest trading date in 1995 ? & \texttt{SELECT account\_id FROM trans WHERE EXTRACT(YEAR FROM "date") = 1995 ORDER BY "date" ASC FETCH FIRST 1 ROWS ONLY} \\

\bottomrule
\end{tabularx}
\caption{Details of the Evaluation Tasks.}
\label{tab:evaltasks}
\end{table*}

\subsection{Ablation Study Evaluation}\label{appendix:abl}
We further provide the detailed results of our ablation study (shown in Figure~\ref{fig:full_ablation_exp}) in Table~\ref{tab:full_ablation_exp}.
\begin{table*}[ht]
\centering
\begin{adjustbox}{width=\textwidth}
\begin{tabular}{lcccccccc}
\toprule
\multirow{2}{*}{\textbf{Model Variant}} & \multicolumn{4}{c}{\textbf{DateTime}} & \multicolumn{2}{c}{\textbf{Financial Analytics}} & \multicolumn{2}{c}{\textbf{\begin{tabular}[c]{@{}c@{}}OracleSQL\\ Compliance (\%)\end{tabular}}} \\ \cmidrule(lr){2-5} \cmidrule(lr){6-7} \cmidrule(lr){8-9} 
 & \textbf{spd-ora} & \textbf{spd-lite} & \textbf{bd-lite} & \textbf{bd-ora} & \textbf{spd+bd-ora} & \textbf{spd+bd-lite} & \textbf{spd-ora} & \textbf{bd-ora} \\ \midrule
\multicolumn{9}{c}{\textbf{Qwen1.5-7B-Chat}} \\ \midrule
OOTB-Baseline & 30.4 & 58.1 & 37.9 & 2.6 & 24.8 & 47.8 & 33.9 & 4.6 \\
SFTed-Setting-B & 49.2 & 60.3 & 45.3 & 14.5 & 73.4 & 77.5 & 62.7 & 18.9 \\
SFTed-Setting-C & 67.4 & 62.6 & 46.6 & 24.4 & 78.7 & 77.8 & 70.1 & 28.0 \\
SFTed-Setting-D & 65.7 & 61.5 & 42.7 & 26.5 & 85.9 & 78.1 & 69.8 & 31.2 \\
SFTed-Setting-E & 72.9 & 60.5 & 44.5 & 29.1 & 85.4 & 79.1 & 66.8 & \textbf{37.0} \\
\textbf{SFTed-Setting-A-Full} & \textbf{74.0} & \textbf{68.7} & \textbf{57.2} & \textbf{33.8} & \textbf{89.5} & \textbf{84.1} & \textbf{77.6} & 34.8 \\ \midrule
\multicolumn{9}{c}{\textbf{Llama3.1-8B-Instruct}} \\ \midrule
OOTB-Baseline & 29.8 & 62.6 & 41.3 & 2.6 & 17.0 & 35.9 & 36.1 & 3.1 \\ 
SFTed-Setting-B & 44.2 & 66.5 & 45.7 & 19.2 & 74.6 & 75.9 & 57.4 & 18.0 \\
SFTed-Setting-C & 61.3 & 63.1 & 47.8 & 33.3 & 76.9 & 76.7 & 72.2 & 30.8 \\
SFTed-Setting-D & 69.1 & 69.3 & 48.9 & \textbf{35.5} & 78.1 & 74.1 & 72.9 & 35.8 \\
SFTed-Setting-E & 68.5 & 62.6 & 50.2 & 34.6 & 80.8 & 75.6 & 73.3 & \textbf{40.8} \\
\textbf{SFTed-Setting-A-Full} & \textbf{81.2} & \textbf{67.6} & \textbf{59.3} & 29.5 & \textbf{83.2} & \textbf{78.2} & \textbf{79.4} & 32.0 \\ \midrule
\multicolumn{9}{c}{\textbf{Mistral-7b-v0.3-Instruct}} \\ \midrule
OOTB-Baseline & 22.1 & 46.4 & 22.2 & 2.6 & 21.1 & 24.5 & 38.4 & 4.4 \\ 
SFTed-Setting-B & 66.3 & 63.1 & 36.5 & 23.9 & 82.6 & 70.7 & 68.4 & 28.8 \\ 
SFTed-Setting-C & 69.6 & 60.9 & 37.4 & 31.2 & 81.3 & 76.6 & 71.2 & 33.7 \\
SFTed-Setting-D & 74.0 & \textbf{70.9} & 37.4 & 35.9 & 82.0 & 79.7 & 71.3 & 35.1 \\
SFTed-Setting-E & \textbf{81.8} & 64.5 & \textbf{39.0} & \textbf{40.6} & 62.0 & 23.8 & 73.7 & \textbf{38.5} \\
\textbf{SFTed-Setting-A-Full} & 74.6 & 65.4 & 38.8 & 31.2 & \textbf{84.5} & \textbf{80.4} & \textbf{77.3} & 28.2 \\ \bottomrule
\end{tabular}
\end{adjustbox}
\caption{\textbf{Performance comparison of model variants on DateTime, Financial Analytics, and OracleSQL tasks for the different distillation scenarios.} Notations: OOTB (Out-Of-The-Box), spd (Spider), bd (Bird), ora (OracleSQL), lite (SQLite).}
\label{tab:full_ablation_exp}
\end{table*}

\begin{table*}[ht]
\centering
\scriptsize 
\resizebox{\textwidth}{!}{ 
\begin{tabular}{ll}
\toprule
                & \textbf{Finetuning Configuration}                           \\ \midrule
\textbf{Pretrained Checkpoints}   & CodeQwen1.5-7B-Chat, Llama3.1-8B-Instruct, Mistral-7B-Instruct-v0.3 \\
\textbf{Batch Size}               & 512 examples per step                    \\
\textbf{Learning Rate}            & 1e-6 (with linear decay)                 \\
\textbf{Warmup Steps}             & 2,000                                   \\
\textbf{Max Sequence Length} & 8192 tokens                         \\
\textbf{Optimizer}                & Paged AdamW 8-bit ($\beta_1=0.9$, $\beta_2=0.95$) \\
\textbf{Weight Decay}             & N/A                                      \\
\textbf{Gradient Clipping}        & 1.0                                      \\
\textbf{Training Steps}           & 20,000                                   \\
\textbf{Evaluation Metrics}       & Checkpoint-based Execution Accuracy      \\
\textbf{Hardware Setup}           & 8 NVIDIA A100 40GB GPUs                  \\ \midrule
& \textbf{Inference Configuration} \\ \midrule
\textbf{Decoding Strategy}        & Random Sampling                          \\
\textbf{Temperature}              & 0.5                                      \\
\textbf{Top-k Sampling}           & 40                                       \\
\textbf{Top-p Sampling} & 0.9                                      \\
\textbf{Max Sequence Length} & 2048 tokens                        \\
\textbf{Batch Size}               & 32                                       \\ \bottomrule
\end{tabular}
}
\caption{Configuration details for training and inference in our experiments.}
\label{tab:hparams_config}
\end{table*}

\section{SQL Dialect Conversion}
We utilize the SQLGlot library \cite{sqlglot} to translate SQL queries from the Bird and Spider datasets from SQLite to the OracleSQL dialect. To enhance the translations, we apply a custom postprocessor to address potential parsing issues and align with OracleSQL conventions.

\section{Prompts in The Distill-C Framework}\label{appendix:prompts}
\subsection{Prompts for NL and SQL Synthesizer Pipelines}
We also present additional prompt templates utilized across various components of our Distill-C framework, including:
\begin{itemize}
\item Figure~\ref{fig:prompt_nl_syn} - An example prompt template for the NL Synthesizer pipeline (AddRef scenario).
\item Figure~\ref{fig:prompt_sql_syn} - An example prompt template for the SQL Synthesizer pipeline (LearnPrior scenario) with a focus on DateTime use case. 
\end{itemize}

\subsection{Prompts for Multi-Step Filtering Pineline}\label{appendix:llm_filtering_prompts}
Given the large scale of the Synthetic SFT Data (over 10,000 instances), manual or human-in-the-loop evaluation is not feasible. 
Therefore, we rely on soft evaluation using multiple strong LLMs as judges, following \cite{verga2024replacingjudgesjuriesevaluating}. 
We employed two primary evaluation phases as shown in Figure~\ref{fig:distillc_multistep_filtering} as follows:
\begin{itemize}
\item \textbf{LLM-based Quality Evaluation.} In this evaluation, each 'judge' LLM assigns a 1-to-5 star score per criterion, with a cut-off as a hyperparameter: consensus on '5 stars' is required for SQL correctness and compliance, and at least '4 stars' for NL quality (Figure~\ref{fig:quality_eval_judge_prompt}). 
\item \textbf{LLM-based Relevance Evaluation} This evaluation step queries multiple LLMs to assess the relevance of a generated example to the use case in the Reference Examples, using prompts in Figure~\ref{fig:relevance_eval_judge_prompt}. Examples marked 'relevant' by all LLMs are added to the final synthetic fine-tuning set, while those marked 'irrelevant' are stored as 'irrelevant examples' for the Input Schema to guide future NL generation (Figure~\ref{fig:distillc_nl_synthesis}).
\end{itemize}

\begin{figure*}[t!]
\centering
\begin{tcolorbox}[
  title=Prompt Example for NL Synthesizer Pipeline (AddRef),
  colframe=black, 
  colback=gray!10,
  boxrule=0.5mm,
  width=\textwidth,
  sharp corners
]

\lstset{frame=none}
\begin{lstlisting}
Given a Input Schema used in a NL2SQL system, your task is to generation 5 new Natural Language queries inspired on the Reference Examples below and appropriate to the Input Schema
- The Reference Examples refer to a particular Customer use-case, which is the target for the data generation
- The new examples should have similar or higher level of complexity of the Reference Examples provided below.
- While taking inspiration on the Reference Examples, you should also be creative in generating original queries.
- In addition to the Reference Example, please refer to Irrelevant Examples (if any) for examples of NL queries that are not relevant to the Customer use-case 
 
## Reference Examples
- show the distance of the flights that arrived before last May
- visits made past more than twelve days
- show a list containing staff names and their respective genders who were assigned 2 days ago
- Find the names of the university which has more faculties in 2002 than every university in Orange county.
- What is all the information about employees hired until June 21, 2002?
 
## Irrelevant Examples
- show oldest ship in the port of Singapore
 
## Input Schema
CREATE TABLE ports(
id INT,
name VARCHAR(255),
country VARCHAR(255)
)
 
CREATE TABLE cargoes(
id INT,
name VARCHAR(255),
tonnage INT,
port_id INT,
load_date DATE
)
 
## Generated Examples
\end{lstlisting}
\end{tcolorbox}
\caption{Prompt Example for NL Synthesizer Pipeline (AddRef).}
\label{fig:prompt_nl_syn}
\end{figure*}

\begin{figure*}[t!]
\centering
\begin{tcolorbox}[
  title=Prompt Example for SQL Synthesizer Pipeline (LearnPrior),
  colframe=black, 
  colback=gray!10,
  boxrule=0.5mm,
  width=\textwidth,
  sharp corners
]

\lstset{frame=none}
{\renewcommand{\baselinestretch}{0.7}\normalsize
\begin{lstlisting}
Here is the database schema context:
 
CREATE TABLE ports(
    id INT,
    name VARCHAR(255),
    country VARCHAR(255)
)
 
CREATE TABLE cargoes(
    id INT,
    name VARCHAR(255),
    tonnage INT,
    port_id INT,
    load_date DATE
)

DateTime Instructions:
- With a DATE_column, refer to the following instructions:
   - 'today': TRUNC(DATE_column) = TRUNC(SYSDATE)
   - 'yesterday': TRUNC(DATE_column) = TRUNC(SYSDATE)-1
   - 'tomorrow': TRUNC(DATE_column) = TRUNC(SYSDATE)+1
   - 'this year': EXTRACT(YEAR FROM DATE_column) = EXTRACT(YEAR FROM SYSDATE)
   - 'this month': TO_CHAR(DATE_column, 'YYYY-MM') = TO_CHAR(SYSDATE, 'YYYY-MM')
   - 'last month': TO_CHAR(DATE_column, 'YYYY-MM') = TO_CHAR(ADD_MONTHS(SYSDATE, -1) 'YYYY-MM')
   - 'next month': TO_CHAR(DATE_column, 'YYYY-MM') = TO_CHAR(ADD_MONTHS(SYSDATE, +1) 'YYYY-MM')
   - 'until last month' TO_CHAR(DATE_column, 'YYYY-MM') <= TO_CHAR(ADD_MONTHS(SYSDATE, -1) 'YYYY-MM')
   - 'until next month' TO_CHAR(DATE_column, 'YYYY-MM') <= TO_CHAR(ADD_MONTHS(SYSDATE, +1) 'YYYY-MM')
   - 'this quarter': TRUNC(DATE_column, 'Q') = TRUNC(SYSDATE, 'Q')
   - 'last quarter': TRUNC(DATE_column, 'Q') = ADD_MONTHS(TRUNC(SYSDATE, 'Q'), -3)
   - 'last year': TO_CHAR(DATE_column, 'YYYY') = TO_CHAR(ADD_MONTHS(SYSDATE, -12), 'YYYY')
   - 'in the last X months': DATE_column BETWEEN ADD_MONTHS(SYSDATE, -X) AND SYSDATE
   - 'in the last X quarters': DATE_column ADD_MONTHS(TRUNC(SYSDATE, 'Q'), -3*X) AND TRUNC(SYSDATE, 'Q')
   - 'in the last X years': DATE_column BETWEEN ADD_MONTHS(SYSDATE, -12*X) AND SYSDATE
   - 'in next X days': (TRUNC(DATE_column) BETWEEN TRUNC(SYSDATE) AND TRUNC(SYSDATE) + X)
   - 'in year XXXX': EXTRACT(YEAR FROM DATE_column) = XXXX
   - 'after year XXXX':  EXTRACT(YEAR FROM DATE_column) > XXXX
   - 'day X of month Y of year Z': TO_CHAR(DATE_column, 'YYYY-MM-DD') = 'ZZZZ-MM-XX' 
   - 'after day X of month Y of year Z':  DATE_column > TO_DATE( 'ZZZZ-YY-XX', 'YYYY-MM-DD')
   - 'next week': TO_CHAR(dueDate, 'YYYY-IW') = TO_CHAR(SYSDATE + 7, 'YYYY-IW')
   - 'in this February: EXTRACT(YEAR FROM DATE_column) = EXTRACT(YEAR FROM SYSDATE) AND EXTRACT(MONTH FROM DATE_column) = 2
   - 'in this October: EXTRACT(YEAR FROM DATE_column) = EXTRACT(YEAR FROM SYSDATE) AND EXTRACT(MONTH FROM DATE_column) = 10
   - 'in last February': EXTRACT(YEAR FROM DATE_column) = EXTRACT(YEAR FROM SYSDATE) - 1 AND EXTRACT(MONTH FROM DATE_column) = 2
   - 'in next February': EXTRACT(YEAR FROM DATE_column) = EXTRACT(YEAR FROM SYSDATE) + 1 AND EXTRACT(MONTH FROM DATE_column) = 2
   - 'from this April':  TRUNC(DATE_column, 'MM') >= ADD_MONTHS(TRUNC(SYSDATE, 'YYYY'), 4-1) # beginning of this year + 3 months to align with start of April (EXTRACT(MONTH not needed here)
   - 'from this January':  TRUNC(DATE_column, 'MM') >= TRUNC(SYSDATE, 'YYYY') # beginning of this year + 0 months to align with start of January (EXTRACT(MONTH not needed here)
   - 'from this October':  TRUNC(DATE_column, 'MM') >= ADD_MONTHS(TRUNC(SYSDATE, 'YYYY'), 10-1) # beginning of this year + 9 months to align with start of October (EXTRACT(MONTH not needed here)
   - 'until this February': TRUNC(DATE_column, 'MM') <= ADD_MONTHS(TRUNC(SYSDATE, 'YYYY'), 2-1) # beginning of this year + 1 months to align with start of February (EXTRACT(MONTH not needed here)
   ... (truncated)
 
Question: What's the total tonnage of all cargoes loaded or unloaded at the port of Singapore before last April
 
Write a SQL query in Oracle SQL dialect, compatible with the latest version of Oracle Database, that answers the question above.
\end{lstlisting}
}
\end{tcolorbox}
\caption{Prompt Example for SQL Synthesizer Pipeline (LearnPrior).}
\label{fig:prompt_sql_syn}
\end{figure*}

\begin{figure*}[t!]
\centering
\begin{tcolorbox}[
  title=LLMs-as-Juries Quality Evaluation Prompt Example,
  colframe=black, 
  colback=gray!10,
  boxrule=0.5mm,
  width=\textwidth,
  sharp corners
]

\lstset{frame=none}
{\renewcommand{\baselinestretch}{0.9}\normalsize
\begin{lstlisting}
Given an input Question and a Oracle SQL query, prepare an assessment based on the following criteria:
 
SQL Correctness
    - Add one star if the Oracle SQL query returns incorrect results
    - Add one more star, i.e. award 2 stars if the Oracle SQL query executes but returns partially correct results
    - Add one more star, i.e. award 2 stars if the Oracle SQL query returns mostly correct results but with minor inaccuracies or omissions
    - Add one more star, i.e. award 2 stars if the Oracle SQL query returns correct results with negligible issues
    - Add one more star, i.e. award 2 stars if the Oracle SQL query returns accurate and complete results as per the requirement
 
Compliance with Oracle SQL Standards
    - Add one star if the SQL query does not follow Oracle SQL standards or best practices, using deprecated or non-standard syntax
    - Add one more star, i.e. award 2 stars if the SQL query loosely follows Oracle SQL standards, with several deviations from best practices.
    - Add one more star, i.e. award 2 stars if the SQL query generally follows Oracle SQL standards but has room for better alignment with best practices.
    - Add one more star, i.e. award 2 stars if the SQL query closely follows Oracle SQL standards and adheres to many best practices.
    - Add one more star, i.e. award 2 stars if the SQL query strictly adheres to Oracle SQL standards and best practices, showcasing exemplary coding standards.
 
Quality of the Natural Language Query
    - Add one star if the natural language query does not match the SQL, or cannot be answered given the provided Schema.
    - Add one more star, i.e. award 2 stars if the natural language query matches the SQL, but the question does not make any sense to be asked (totally unrealistic).
    - Add one more star, i.e. award 3 stars if the natural language query is consistent with the SQL, but it it does not look natural (no domain knowledge, the style looks synthetic-templated, does not use "domain-specific" words).
    - Add one more star, i.e. award 4 stars if the natural language query is correct and consistent, but the NL Question can further be improved for clarity, conciseness, small typos.
    - Add one more star, i.e. award % stars if the natural language query is perfect.
 
The Schema context is provided below.
 
CREATE TABLE ports(
    id INT,
    name VARCHAR(255),
    country VARCHAR(255)
)
 
CREATE TABLE cargoes(
    id INT,
    name VARCHAR(255),
    tonnage INT,
    port_id INT,
    load_date DATE
)
 
Question: What's the total tonnage of all cargoes loaded or unloaded at the port of Singapore before last April
Oracle SQL: SELECT SUM(c.tonnage) FROM cargoes c JOIN ports p ON c.port_id = p.id WHERE p.name = 'Singapore' AND (EXTRACT(YEAR FROM c.load_date) < EXTRACT(YEAR FROM SYSDATE) - 1 OR (EXTRACT(YEAR FROM c.load_date) EXTRACT(YEAR FROM SYSDATE) - 1 AND EXTRACT(MONTH FROM c.load_date) < 4));
 
The output must have following items in an orderly manner:
- The final star ratings of criterions in a list-wise manner
- The final star ratings of criterions in a json format
- Explain the scores with a short text (< 100 words).
\end{lstlisting}
}
\end{tcolorbox}
\caption{Prompt for LLM-based Quality Evaluation.}
\label{fig:quality_eval_judge_prompt}
\end{figure*}

\begin{figure*}[t!]
\centering
\begin{tcolorbox}[
  title=LLMs-as-Juries Relevance Evaluation Prompt Example,
  colframe=black, 
  colback=gray!10,
  boxrule=0.5mm,
  width=\textwidth,
  sharp corners
]

\lstset{frame=none}
\begin{lstlisting}
Given an a Natural Language query and the corresponding SQL Query generated for a NL2SQL Model, your goal is to assess whether the generated example is relevant to the Customer use-case represented by any of the Reference Examples shown below.
 
## Reference Examples
- show the distance of the flights that arrived before last May
- visits made past more than twelve days
- show a list containing staff names and their respective genders who were assigned 2 days ago
- Find the names of the university which has more faculties in 2002 than every university in Orange county.
- What is all the information about employees hired until June 21, 2002?
- Show me the aircraft names that travelled 8430 kms that departed before November of 4 years ago
- How many students exist who are registered with just a single allergy?
- show all maintenance contracts that end until next Dec
- Give me the list of actors which was last updated until last Saturday
- show the distance of the flights that arrived before last January
- show all machines made in 1992
- Show me invoices that are due to be paid in the next half year.
- What is all the information about employees hired until June 21, 2002?
- show all order items delivered before last march
- Show the number of attendees in year 2008 or 2010.
- Show me all students who registered for a course from 3 days ago, including the course name and student details.
- give people addresses who lived on address till april.
- List all customers who placed an order from the next 30 days and the order status is 'New'.
- show all maintenance contracts that end until next May
- How many customers are not responded to mailshot sent from week 5 2018
 
## Input Natural Language query and SQL query
Natural language query: What's the total tonnage of all cargoes loaded or unloaded at the port of Singapore before last April
SQL Query: SELECT SUM(c.tonnage) FROM cargoes c JOIN ports p ON c.port_id = p.id WHERE p.name = 'Singapore' AND (EXTRACT(YEAR FROM c.load_date) < EXTRACT(YEAR FROM SYSDATE) - 1 OR (EXTRACT(YEAR FROM c.load_date) EXTRACT(YEAR FROM SYSDATE) - 1 AND EXTRACT(MONTH FROM c.load_date) < 4));
 
## Assessment ("**Relevant**"/"**Irrelevant**")
\end{lstlisting}

\end{tcolorbox}
\caption{Prompt for LLM-based Relevance Evaluation.}
\label{fig:relevance_eval_judge_prompt}
\end{figure*}
